\documentclass{article}

\usepackage[final]{nips_2017}

\usepackage[utf8]{inputenc} 
\usepackage[T1]{fontenc}    
\usepackage{hyperref}       
\usepackage{url}            
\usepackage{booktabs}       
\usepackage{amsfonts}       
\usepackage{nicefrac}       
\usepackage{microtype}      

\usepackage{amsmath}
\usepackage{amssymb}

\usepackage[linesnumbered,algoruled,boxed,lined]{algorithm2e}

\makeatletter 
\g@addto@macro{\@algocf@init}{\SetKwInOut{Parameter}{Parameters}} 
\g@addto@macro{\@algocf@init}{\SetKwInOut{Initialize}{Initialize}} 
\makeatother

\usepackage{graphicx}
\usepackage{sidecap} 

\usepackage{wasysym} 
\usepackage{enumitem}
\setlist[itemize]{noitemsep, topsep=0pt}

\usepackage{color}
\usepackage{colortbl}
\usepackage[table]{xcolor}

\def\EWGt{{\mbox{$p_t$}}}
\def\RWGt{{\mbox{r$_t$}}}
\def\RWGtm{{\mbox{r$_{t-1}$}}}
\def\RWG1{{\mbox{r$_{t+1}$}}}

\def\NRGWt{\|\RWGt\|^2}

\def\zti{\mbox{r$_{t,i}^2$}}

\def\XX{{\sc SALeRA}}
\def\bXX{{\sc {\bf SALeRA}}}
\def\pXX{{\sc SPALeRA}}
\def\bpXX{{\sc {\bf SPALeRA}}}
\def\beve{{\sc {\bf ALeRA}}}
\def\eve{{\sc ALeRA}}

\def\adam{{\sc Adam}}
\def\Adam{{\sc Adam}}
\def\bAdam{{\sc {\bf Adam}}}
\def\NAG{{\sc NAG}}
\def\bNAG{{\sc {\bf NAG}}}
\def\Adagrad{{\sc Adagrad}}
\def\bAdagrad{{\sc {\bf Adagrad}}}
\def\Adadelta{{\sc Adadelta}}
\def\adameve{{\sc Ag-Adam}}
\def\badameve{{\sc {\bf Ag-Adam}}}

\def\RR{{\rm I\hspace{-0.50ex}R}}
\def\EE{{\mathbb{E}}} 

\title{Stochastic Gradient Descent: Going As Fast As Possible But Not Faster}

\author{
  Alice Schoenauer Sebag \\
  Altschuler\&Wu lab, Dpt of Pharm. Chem.\\
  UCSF, San Francisco, CA 94158 \\
  \texttt{alice.schoenauersebag@ucsf.edu} \\
  \And
  Marc Schoenauer \\
  INRIA-CNRS-UPSud-UPSay \\
  TAU, U. Paris-Sud, 91405 Orsay\\
  \texttt{marc.schoenauer@inria.fr} \\
  \And
  Mich\`ele Sebag \\
  CNRS-UPSud-INRIA-UPSay \\
  TAU, U. Paris-Sud, 91405 Orsay\\
  \texttt{michele.sebag@lri.fr} \\
}

\begin{document}
\maketitle

\begin{abstract}
 When applied to training deep neural networks, stochastic gradient descent (SGD) often incurs steady progression phases, interrupted by catastrophic episodes in which loss and gradient norm explode. A possible mitigation of such events is to slow down the learning process. 
 
 This paper presents a novel approach to control the SGD learning rate, that uses two statistical tests. The first one, aimed at fast learning, compares the momentum of the normalized gradient vectors to that of random unit vectors and accordingly gracefully increases or decreases  the learning rate. The second one is a change point detection test, aimed at the detection of catastrophic learning episodes; upon its triggering the learning rate is instantly halved. \\
 Both abilities of speeding up and slowing down the learning rate allows the proposed approach, called \XX, to learn as fast as possible but not faster. Experiments on standard benchmarks show that \XX\ performs well in practice, and compares favorably to the state of the art. 

 

\end{abstract}

Machine Learning (ML) algorithms require efficient optimization techniques, whether to solve convex problems (e.g., for SVMs), or non-convex ones (e.g., for Neural Networks). In the convex setting, the main focus is on the order of the convergence rate \citep{nag,SAGA}. In the non-convex case, ML is still more of an experimental science. Significant efforts are devoted to devising optimization algorithms (and robust default values for the associated hyper-parameters)  
tailored to the typical regime of ML models and problem instances (e.g. deep convolutional neural networks for MNIST~\citep{mnist} or ImageNet~\citep{ImageNet_VSS09})~\citep{adagrad,adadelta,pesky,adam,rmsprop}.

As the data size and the model dimensionality increase, mainstream convex optimization methods are adversely affected. 
Hessian-based approaches, which optimally handle convex optimization problems however ill-conditioned they are, do not scale up and approximations are required~\citep{Approx-HessianICML12}. Overall, Stochastic Gradient Descent (SGD) is increasingly adopted both in convex and non-convex settings, with good performances and linear tractability ~\citep{bottou2008,Singer2015}.

Within the SGD framework, one of the main issues is to know how to control the learning rate: the objective is to reach a satisfactory learning speed without triggering any catastrophic event, manifested by the sudden rocketing of the training loss and the gradient norm. Finding "how much is not too much" in terms of learning rate is a slippery game. It depends both on the current state of the system (the weight vector) and the current mini-batch. Often, the eventual convergence of SGD is ensured by decaying the learning rate as $O(t)$ \citep{RobbinsMonro:1951,SAGA} or $O(\sqrt{t})$ \citep{Zinkevich:2003} with the number $t$ of mini-batches. While learning rate decay effectively prevents catastrophic events, it is a main cause for the days or weeks of computation behind the many breakthroughs of deep learning. Many and diverse approaches have thus been designed to achieve the learning rate adaptation \citep{Amari,adagrad,pesky,adam,rmsprop,L2L} (more in Section~\ref{sec:soa}).

This paper proposes a novel approach to adaptive SGD, called \XX\ ({\em Safe Agnostic LEraning Rate Adaptation}). \XX\ is based on the conjecture that, if learning catastrophes are well taken care of, the learning process can speed up whenever successive gradient directions show general agreement about the direction to go.
\\
The frequent advent of catastrophic episodes, long observed by neural net practitioners~\citep[Chapter~ 8]{DLBook} raises the question of how to best mitigate their impact. The answer depends on whether these events could be anticipated with some precision. 
Framing catastrophic episodes as random events\footnote{As far as tractable computational resources are involved in their prediction.}, we adopt a purely curative strategy (as opposed to a preventive one): detecting and instantly curing catastrophic episodes. Formally, a sequential cumulative sum change detection test, the Page-Hinkley (PH) test \citep{Page,Hinkley} is adapted and used to monitor the learning curve reporting the minibatch losses. If a change in the learning curve is detected, the system undergoes an instant cure by halving the learning rate and backtracking to its former state. Such instant cure can be thought of in terms of a dichotomic approximation of line search (see e.g. \cite{SAGA}, Eq. (3)).\\
Once the risk of catastrophic episodes is well addressed, the learning rate can be adapted in a more agile manner: the \eve\ ({\em Agnostic LEarning Rate Adaptation}) process increases (resp. decreases)
the learning rate whenever the correlation among successive gradient directions is higher (resp. lower) than random, by comparing the actual gradient momentum and the agnostic momentum built from random unit vectors.

The contribution of the paper is twofold. First, it proposes an original and efficient way to control learning dynamics (section \ref{sec:speedup}). Secondly, it opens a new approach for handling catastrophic events and salvaging a significant part of doomed-to-fail runs (section \ref{sec:PH}). The experimental validation thereof compares favorably with the state of the art on the MNIST and CIFAR-10 benchmarks (section \ref{experiments}).

\section{Related work} \label{background}\label{sec:soa}

SGD was revived in the last decade as an effective method for training deep neural networks with linear computational complexity in the size of the dataset \citep{bottou2008,Singer2015}.
SGD faces two limitations, depending on the learning rate: too large, and the learning trajectory leads to catastrophic episodes; too small, and its convergence takes ages. The dynamic adjustment of the learning rate was therefore acknowledged a key issue since the early SGD days ~\citep{RobbinsMonro:1951}.
 

\paragraph{Dealing with catastrophic events in deep learning} The exploding gradient problem, described in~\cite[Chapter~8]{DLBook} as the encounter of steep cliff structures in the derivative landscape during learning, is frequently met while training neural networks (and even more so when training recurrent neural networks 
\citep{Bengio:Exploding-gradients1}). \\
When it comes to dealing with such events, most published work focuses on creating the conditions so that they do not arise. Among the possibilities are the use of regularizations, e.g., $L_1$ or $L_2$ regularization~\citep{Bengio:Exploding-gradients2} or Max-norm regularization~\citep{dropout}. Gradient clipping, constraining the gradient norm to remain smaller than a constant~\citep{Bengio:Exploding-gradients2}, is another possibility. The introduction of batch normalization~\citep{batchnormalization}
also helps diminishing the frequency of such events. Finally, proper initialization~\citep{Glorot2010,nesterovmomentum} or unsupervised pre-training~\citep{Erhan:2010}, i.e. initializing the optimization trajectory in a good region of the parameter space,  also diminish the frequency of such events. 

\paragraph{Learning rate adaptation} Addressing the slow speed of SGD, learning rate adaptation has been acknowledged a key issue since the late 80s (see~\cite{George:2006} for a review). Using the information contained in the correlation of successive gradient directions was already at the heart of the delta-delta and delta-bar-delta update rules proposed by \cite{Jacobs:1988}. Briefly, the delta-bar-delta rule states that for each parameter, if the current gradient and the relaxed sum of past gradients have the same sign, the learning rate is incremented additively; and if they are of opposite sign, then the learning rate is decremented multiplicatively. Decrementing the learning rates faster than increasing them was already advocated by the author to adapt faster in case of catastrophic events.\\
 The natural gradient descent (NGD) approach~\citep{Amari} considers the Riemaniann geometry of the parameter space, using the Fisher information matrix (estimated by the gradient covariance matrix), to precondition the gradient. Due to its quadratic complexity in the dimension of the parameter space, NGD approximations have been designed for deep networks~\citep{Pascanu:2014}. Notably, approaches such as the Hessian-Free from \cite{HessianFreeOptimization} can be interpreted as NGD~\citep{Pascanu:2014}.\\
\Adagrad~\citep{adagrad} also uses the information of past gradients to precondition each update in a parameter-wise manner, dividing the learning rate by the sum of squared past gradients. 
Several approaches have been proposed to refine \Adagrad\ and mitigate its learning rate decay to 0, including 
 \Adadelta~\citep{adadelta}, RMSProp~\citep{rmsprop}, and \Adam~\citep{adam}. \Adam\ is based on estimating the first and second moments of the gradient w.r.t. each parameter, and using their ratio to update the parameters. 
Moment estimates are maintained by exponential moving averages of different weight factors, such that by default the inertia of the first moment is higher by two orders of magnitude than the second.
As will be seen, \XX\ also builds upon the use of the gradient second moment, with the difference that it is compared with a fixed agnostic counterpart.

In \citep{pesky}, the learning rate is computed at each time-step to approximately maximally decrease the expected loss, where the loss function is locally approximated by a parabola. Finally,~\cite{L2L} address learning rate adaptation as a reinforcement learning problem, exploiting the evidence gathered in the current time steps to infer what would have been the good decisions earlier on, and accordingly optimizing a hyper-parameter adjustment policy.

More remotely related are the momentum approaches, both its classic~\citep{classicmomentum} and Nesterov versions~\citep{nag} as derived by~\cite{nesterovmomentum}, that rely on a relaxed sum of past gradients for indicating a more robust descent direction than the current gradient.


\section{\XX} \label{eve}
\XX\ involves two components: a learning rate adaptation scheme, which ensures that the learning system goes as fast as it can; and a catastrophic event manager, which is in charge of detecting undesirable behaviors and getting the system back on track.
\subsection{Agnostic learning rate adaptation}\label{sec:speedup}
\paragraph{Rationale} The basic idea of the proposed learning rate update is to compare the current gradient descent to a random walk with uniformly chosen gradient directions. Indeed, the sum of successive normalized gradient vectors, referred to as cumulative path in the following, has a larger norm than the sum of uniformly drawn unit vectors if and only if gradient directions are positively correlated. In such cases, the learning process has a global direction and the process can afford to speed up. On the opposite, if the norm of the cumulative path is smaller than its random equivalent, gradient directions are anti-correlated: the process is alternating between opposite directions (e.g., bouncing on the sides of a narrow valley, or hovering around some local optimum) and the learning rate should be decreased.

The \eve\ scheme takes inspiration from the famed CMA-ES \citep{hansenECJ01} and NES \citep{NES-JMLR14} algorithms, today considered among the best-performing derivative-free continuous optimization algorithms. These approaches de facto implement natural gradient optimization \citep{Amari} and instantiate the Information-Geometric Optimization paradigm \citep{IGO} in the space of normal distributions on $\RR^d$. 
Formally, CMA-ES maintains a normal distribution. The variance of the normal distribution, aka {\em step-size}, is updated on the basis of a comparison of the cumulative path of the algorithm (moving exponential average of successive steps) with that of a random walk with Gaussian moves of fixed step-sizes. This mechanism is said to be agnostic as it makes no assumption whatsoever on the properties of the optimization objective.

\paragraph{The \eve\ algorithm} The partial 
adaptation of the CMA-ES scheme to the minimization of a loss function $\cal L$ on a d-dimensional parameter space is defined as follows.  Let $\theta_t$ be the solution at time $t$, $g_t = \nabla_\theta {\cal L}(\theta_t)$ the gradient of the current loss, 
and $\eta_t$ the current learning rate. SGD computes the solution at time $t+1$ by $\theta_{t+1} = \theta_t - \eta_t g_t $. Let $\|.\|$ denote the $L_2$ norm and $<.,.>$ the associated dot product. 

{\bf Definition. } \textit{ 
For $t>0$, the exponential moving average of the normalized gradients with weight $\alpha$, \EWGt, and its random equivalent \RWGt\ are defined as:}
\begin{align}
p_0&=0; ~~\mbox{r}_0=0\\
p_t &=  \alpha \, g_t/\lVert g_t \rVert + (1-\alpha) \, p_{t-1}\\
\RWGt &= \alpha \, u_t + (1-\alpha) \, \RWGtm
\end{align}
\textit{where $(u_t)_{t>0}$ are independent random unit vectors in $\RR^d$.}

{\bf Proposition. } \textit{For $t>0$, the expectation $\mu(t)$ and variance $\sigma^2(d,t)$ of $\NRGWt$ as defined above are:}
\begin{equation}\label{random_mean_t} 
	\mu(t) = \mathbb{E}(\|\RWGt\|^2) = \dfrac{\alpha}{2-\alpha}[1-(1-\alpha)^{2t}] 
\end{equation}
\begin{equation}\label{random_var_t}
    \sigma^2 (d, t) = Var(\NRGWt) = \dfrac{1}{d}\frac{2\alpha^2(1-\alpha)^2}{(2-\alpha)^2((1-\alpha)^2+1)} [1-(1-\alpha)^t] [1-(1-\alpha)^{t-1}] 
\end{equation}
{\em Proof}:  Appendix~A.
%

\def\minib{$\n_m$}
\paragraph{} 
Let $\mu$ and $\sigma^2(d)$ denote the limits of respectively $\mu(t)$ and $\sigma^2 (d, t)$ as $t\rightarrow+\infty$. 
At time step $t$, the \eve\ scheme updates the  cumulative path \EWGt\ 
by comparing its norm with the distribution of the agnostic momentum defined from $\mu$ and $\sigma^2(d)$. 
The learning rate is increased or decreased depending on the normalized gap between the squared norm of $p_t$ and \RWGt:  
\begin{equation}\label{updateeq}
\eta_{t+1} = \eta_t \exp \left( C \dfrac{\lVert p_t \rVert^2 - \mu}{\sigma(d)}  \right)
\end{equation}
with $C$ a hyper-parameter of the algorithm.

\paragraph{} The approach is implemented in the \eve\ algorithm (non-greyed lines in Algorithm \ref{alg.mainLoop}). Given hyper-parameters $\alpha$ and $C$, as well as $\eta_0$, the initial learning rate, and $\rho$ the mini-batch size, each iteration over a mini-batch computes the new exponential moving average of the normalized gradient (line \ref{movingAverage}), performs the agnostic update of the learning rate (line \ref{updateLearningRate}) before updating parameter $\theta$ the usual way (line \ref{standardParameterUpdate}).
The learning rate can be controlled in a layer-wise fashion, independently maintaining an exponentiated moving average and updating the learning rate for each layer of the neural network (lines \ref{movingAverage}-\ref{updateLearningRate}). This algorithm is used in all experiments of Section \ref{experiments}.

\paragraph{Parameter-wise learning rate adaptation}\label{pagma} As noted by \cite{adam}, the parameter-wise control of the learning rate is desirable in some contexts. The above scheme is extended to achieve the parameter-wise update of the learning rate as follows. 
\def\qti{{$\mbox{r}_{t,i}^2$}}
For $t>0$ let \qti\ denote the squared i$^{th}$ coordinate of \RWGt. It is straightforward to show that the expectation (respectively the standard deviation) of \qti\ is the expectation of \RWGt\ divided by $d$ (respectively the standard deviation of \RWGt\ divided by $\sqrt{d}$).

Given $t>0$ and $i \in [|1;d|]$, the squared i$^{th}$ coordinate of $p_{t}$ noted \zti, can thus likewise be adjusted by comparison with its random counterpart \qti. The update therefore becomes:
\begin{equation}
\eta_{i,t+1} = \eta_{i,t} \exp \left\{ C' 
\frac{1}{\sqrt{d}} 
\frac{\zti - \mu}{\sigma(d)}
\right\}
\end{equation}
See Appendix B for a full derivation of the parameter-wise algorithm.

\definecolor{shadecolor}{RGB}{200,200,200}
\newcommand{\PH}[1]{\colorbox{shadecolor}{#1}}

\begin{algorithm}[tb]
\KwIn{~Model with loss function $\cal L$}
\Parameter{~Memory rate $\alpha$, factor $C$ \hfill //algorithm parameters \label{algo-parameters}\\
Initial learning rate $\eta_0$, mini-batch ratio $\rho$
\hfill // run parameters \label{run-parameters}
}
\Initialize{~$t\leftarrow 0; p\leftarrow 0; \eta\leftarrow \eta_0$; $\mbox{init} (\theta)$ \hfill // initialization \\
\PH{$L,L_{min}, \ell, \bar{\ell}, \leftarrow 0$; 
$\Delta \leftarrow {\cal L}(\theta, \mbox{first mini-batch})/10$
}  \hfill // initialize Page-Hinkley 
}
\While{\text{stopping criterion not met}}{
$MB \leftarrow \mbox{new mini-batch}$ ; $t\leftarrow t+1$ \hfill // perform forward pass \\
$\ell = \rho {\cal L}(\theta, \mbox{MB}) + (1-\rho) \ell$\\
\PH{$\bar{\ell} \leftarrow  (\ell + t \bar{\ell})/(t+1)$} \hfill //  empirical mean of batch losses \label{movingLossAverage}\\
\PH{$L \leftarrow L + (\ell-\bar{\ell})$} \hfill // cumulated deviations from mean \label{cumulatedDeviations}\\ 
\PH{$L_{min} \leftarrow min (L_{min}, L)$} \hfill // lower bound of deviations \label{lowerBound}\\
\uIf{\PH{$L - L_{min} > \Delta$}}{
\PH{$\theta \leftarrow \theta^{(b)}$; $\eta \leftarrow \eta/2$ } \hfill // Page-Hinkley triggered: backtrack \label{backtrack}\\
\PH{$L, L_{min}, \bar{\ell}, \ell \leftarrow 0$; $t \leftarrow 0$} \hfill // and re-initialize Page-Hinkley\footnotemark \label{reinitialisation}
}
\Else{
\PH{$\theta^{(b)} \leftarrow \theta$} \hfill // save for possible backtracks\\
$g \leftarrow \nabla_{\theta}{\cal L}(\theta, \mbox{MB})$ \hfill // compute gradient with backward pass\\
$p \leftarrow \alpha g/\lVert g \rVert + (1-\alpha) p$ \hfill // exponential moving average of normalized gradients  \label{movingAverage}\\
$\eta \leftarrow \eta \exp \left( C ( \lVert p \rVert^2_2 - \mu) / \sigma(d) \right)$ \hfill // agnostic learning rate update \label{updateLearningRate}\\
$\theta \leftarrow \theta - \eta g$ \hfill // standard parameter update \label{standardParameterUpdate}
}

}
\caption{\PH{S}\hspace{-0.15cm}-\eve: {\em Agnostic LEarning Rate Adaptation} and \PH{Page-Hinkley change detection}}\label{alg.mainLoop}
\end{algorithm}

\subsection{Catastrophic event management}\label{sec:PH}
As said, the ability to learn fast requires an emergency procedure, able both to detect an emergency and to recover from it. 
\def\meta{\eta^{-}}

\paragraph{Recovery}
\footnotetext{There are in fact two time counters in the full \XX\ Algorithm, the global usual one and the one related to the Page-Hinkley test. Only one is used here for simplicity reasons.}


In a healthy learning regime, the training error should decrease along time $-$ up to the noise due to the inter-batch variance $-$ 
unless the learning system abruptly meets a cliff structure \citep{DLBook}, usually blamed on too large a learning rate in an uneven gradient landscape.\\
In a convex noiseless optimization setting, if computationally tractable, the best strategy is to compute (an approximation of) the optimal learning rate $\eta^*$ through line search\footnote{See \cite{SAGA} on how to handle the noise in the case of a composite loss function.}. In such context, as a thought experiment, let $\meta > 0$ be such that, if used to update $\theta_t$, the resulting $\theta_{t+1}$ would yield the same performance as $\theta_t$. For $\eta > \meta$,  $\theta_{t+1}$ yields a {\em worse} performance than $\theta_t$, and if continued the optimization process is likely to diverge. For $\eta < \eta^*$, $\theta_{t+1}$ yields a performance improvement. For 
$\eta^* < \eta < \meta$,  $\theta_{t+1}$ yields a performance improvement too; the further $\theta$ trajectory is likely to bounce back and forth on the walls of the optimum valley.

Overall, the safety zone for the learning rate is $]0, \meta[$ (with the caveat that the safety zone is narrower for ill-conditioned optimization problems). The proposed safeguard strategy primarily aims to detect when $\eta$ steps outside of its safety zone 
(see the change point detection test below), and to apply a correction as to get back in it. Upon change detection in the mini-batch loss, \XX\ implements a straightforward correction: halving the learning rate and recovering the last solution before 
test triggering. The halving process is iterated if needed, sending $\eta_t$ back in $]\frac{\meta}{2}, \meta]$ exponentially fast (except for the perturbations in the gradient due to the mini-batch variance).\\ 
The rationale for the halving trick is based on a trade-off between the number of successive dividing iterations, that could indeed be made even smaller by using a larger dividing factor, and the required standard \eve\ iterations that will be needed to reach the optimal learning rate $\eta^*$ after having reached the safety zone again. The choice of the $2$ dividing factor is further discussed in Appendix C.

\paragraph{Detection} \XX\ applies a change detection test on the signal given as the minibatch loss $\ell_t$. 
The PH detection test \citep{Page,Hinkley} is chosen as it provides optimal guarantees about the trade-off between the detection delay upon a change (affecting the average or standard deviation of the signal) and the mean time between false alarms. For $t>0$, it maintains the empirical mean $\bar{\ell_t}$ of the signal, and the cumulative deviation\footnote{The PH test takes into account the extreme value phenomenon by considering an upper cumulative deviation and a lower cumulative deviation, defined from $L_t$ by adding (resp. subtracting) a margin $\delta$ to $L_t$. In \XX\ $\delta$ is set to 0.} $L_t$ from the empirical mean ($L_t = L_{t-1} + (\ell_t - \bar{\ell_t})$). Finally, it records the empirical bounds of $L_t$ ($L_{min} = min_t L_t$; $L_{max} = max_t L_t$). In case of a stationary signal, the expectation of $L_t$ is 0 by construction; the PH change test is thus triggered when the gap between $L_t$ and its empirical bounds is higher than a problem-dependent threshold $\Delta$, which controls the alarm rate. \\
The PH test is implemented in the \XX\ algorithm as follows (greyish lines in Algorithm \ref{alg.mainLoop}). $\Delta$ is set to ${\cal L}(\theta, \mbox{first mini-batch})/\lambda$, with $\lambda=10$, that is, one tenth of the empirical loss on the first minibatch in all experiments (this issue is further discussed in   Section~\ref{sec:experimental_results}). Variables $\bar{\ell_t}$, $L$ and $L_{max}$ are maintained (lines \ref{cumulatedDeviations}- \ref{lowerBound}). In the learning context, a decrease of the loss signal is welcomed and expected. Only the case of an increasing signal is thus monitored. Upon test triggering ($L_{max} - L > \Delta$), the learning rate is halved and the weight vector $\theta$ is reset to the last solution before then (line \ref{backtrack}), and the PH test is reinitialized (line  \ref{reinitialisation}).

\section{Experiments} \label{experiments}

\newcommand{\Agnostic}{agnostic}

The goal of the following experiments is to validate the algorithmic ideas introduced in section \ref{eve} by comparing their application with that of widely used optimization techniques (see Section \ref{background}) on some straightforward NN architectures. 

\subsection{Experimental Setting}
\label{sec:parameters}
\paragraph{Datasets} All experiments are performed on the {\bf MNIST}~\citep{mnist} and {\bf CIFAR-10}~\citep{cifar} datasets, which respectively contain 60k and 50k training examples. Both contain 10k test examples, which are to be classified in 10 classes. The data is normalized according to the mean and standard deviation along each coordinate on the training set.  

\paragraph{Algorithms} \bAdagrad, \bNAG, and \bAdam\ are used as baselines. The \Agnostic\ adaptation rule and the change detection can be applied independently. In order to separate their effect, 3 original algorithms are studied here: \beve\ (the white lines in Algorithm \ref{alg.mainLoop}) implements the \Agnostic\ learning rate adaptation without the change detection; \badameve\ uses the same \Agnostic\ adaptation for the learning rate on top of the {\Adam} algorithm. Finally, the change detection mechanism is implemented with the \Agnostic\ adaptation, yielding \bXX~ as described in Algorithm \ref{alg.mainLoop}, as well as its parameter-wise version \bpXX\ (Algorithm \ref{pAlgo} in Appendix B).

\paragraph{Hyperparameters} The exploration of the hyper-parameter space for all algorithms has been done on a grid of possible values (with the exception of \bAdagrad\ which has no hyper-parameter):
\begin{itemize}[noitemsep,topsep=0pt]
\item[-] \bNAG: the momentum $\gamma \in \{.8, .9, .99, .999\}$.
\item[-] \bAdam: $\beta1 \in \{.8, .9, .99\}$ and $\beta2 \in \{.99, .999, .9999\}$;
\item[-] \beve: $\alpha \in \{.001, .01, .1, .25\}$ and $C \in \{3.10^{-8}, 3.10^{-7}, 3.10^{-6}, 3.10^{-5}\}$ (see Algorithm \ref{alg.mainLoop}). The parameters for \bXX\ and \bpXX\ are the same, as there is no additional parameter for the Page-Hinkley part.
\item[-] \badameve: the recommended values for \Adam\ ($\beta1=.9$ and $\beta2=.999$) are used for the \Adam\ part, the same values than for \eve\ are used for the \eve\ part.
\end{itemize} 
The initial learning rate $\eta_0$ ranges from $10^{-5}$ to $1$ depending on the algorithm. Finally, the mini-batch size was set to either 1\% or 1\permil\ of the training set size. All reported results are based on 5 independent runs performed for each hyperparameter set unless otherwise specified.

\paragraph{Network Models} All experiments consider the following 4 network architectures:
\begin{itemize}
\item[-] {\bf M0}: a softmax regression model with cross-entropy loss (i.e. no hidden layers),
\item[-] {\bf  M2}: 2 fully connected hidden layers with ReLU activation, on top of M0. The hidden layers are of respective sizes (500, 300) for MNIST, (1~500, 900) for CIFAR-10. 
\item[-] {\bf  M2b} is identical to Model M2 above, except that Batch Normalization layers~\citep{batchnormalization} are added in each hidden layer.
\item[-] {\bf  M4}: LeNet5-inspired convolutional models \citep{mnist}. These models contain 2 convolutional layers with max-pooling followed by 2 fully connected layers, all with ReLU activation. They are of respective sizes (32, 64, 128, 128) for MNIST, and (32, 64, 384, 384) for CIFAR-10. Batch normalization is used in each layer.
\end{itemize}
These architectures are not specifically optimized for the task at hand, but rather chosen to compare the performances of past and novel algorithms in a wide variety of situations.

\paragraph{Experimental conditions} All computations are performed on 46 GPUs (5 TITAN X(Pascal), 9 GTX 1080 and 32 Tesla K80) using the Torch library \citep{torch} in double precision. A typical run on a TITAN X (Pascal) GPU for 20k mini-batches of size 50 for CIFAR-10 on M4 takes between 8 and 10 minutes for all algorithms. 

\paragraph{Metrics} MNIST and CIFAR-10 are classification problems. We therefore report classification accuracy on the test set at 5 epochs (i.e. 5 full passes on the training set) and 20 epochs (end of all runs), as well as their standard deviations on the 5 independent runs.

\subsection{Experimental results}
\label{sec:experimental_results}
\setlength{\tabcolsep}{4pt}
\setlength{\tabcolsep}{4pt}
\begin{table}[htbp]
{\footnotesize
	{\centering
\begin{tabular}{l | l | l || c c c | c c c c |} \hline
&  & & \NAG & Adagrad & \adam & \adameve & \eve & \XX &\pXX \\ \hline
\tiny{MNIST} & M0 & 5ep. & 7.75 (.14) & 7.73 (.06) & 7.72 (.12) & \textbf{7.31 (.09)} & 7.51 (.09) & 7.51 (.09) & 8.03 (.09)\\ 
&  & 20ep. & 7.59 (.07) & 7.51 (.09) & 7.43 (.10) & \textbf{7.29 (.09)} & 7.43 (.03) & 7.44 (.04) & 7.60 (.08)\\ 
\hline
 & M2 & 5ep. & \textit{1.95 (.17)} & 2.00 (.06) & 2.07 (.11) & \textit{1.93 (.17)} & \textbf{1.86 (.11)} & 1.87 (.05) & \textit{1.93 (.14)}\\ 
 &  & 20ep. & \textit{1.58 (.08)} & 1.71 (.08) & \textit{1.56 (.06)} & \textit{1.57 (.04)} & \textbf{1.55 (.10)} & \textit{1.59 (.09)} & \textbf{1.55 (.09)}\\ 
 \hline
 & M2b & 5ep. & 1.82 (.13) & 1.72 (.07) & 1.81 (.07) & \textit{1.66 (.08)} & \textbf{1.59 (.08)} & \textbf{1.59 (.08)} & 1.78 (.08)\\ 
 &  & 20ep. & \textit{1.47 (.10) }& 1.48 (.06) & 1.57 (.94) & 1.53 (.05) & \textbf{1.43 (.04)} & 1.48 (.09) & 1.50 (.09)\\ 
 \hline
 & M4b & 5ep. & \textit{.85 (.08)} & 1.02 (.09) & \textit{.89 (.31)} & \textit{.91 (.06)} & .\textbf{82 (.30)} & .\textbf{82 (.30)} & \textbf{.82 (.14)}\\ 
 &  & 20ep. & .72 (.09) & .82 (.08) & .80 (.08) & .79 (.05) & \textbf{.63 (.05)} & \textit{.64 (.07)} & \textbf{.63 (.11)}\\ 
\hline 
 \hline
\tiny{CIFAR} & M0 & 5ep. & 60.37 (.55) & 60.49 (.71) & 60.60 (.45) & \textbf{59.62 (.27)} & \textit{59.89 (.19)} & \textit{59.69 (.33)} & 61.32 (.65)\\ 
 & M0 & 20ep. & 59.73 (.19) & 59.76 (.36) & 59.81 (.24) & \textit{59.34 (.24)} & \textbf{59.31 (.11)} & \textbf{59.31 (.25)} & 59.71 (.48)\\ 
 \hline
 & M2 & 5ep. & 45.82 (.93) & 44.81 (.62) & 45.68 (.39) & 44.91 (.42) & 44.69 (17.58) & \textbf{44.42 (.24)} & 45.74 (.85)\\ 
 & M2 & 20ep. & 45.08 (.32) & 43.59 (.51) & 44.43 (.50) & 43.25 (.40) & 43.19 (.21) & \textbf{42.72 (.41)} & 44.48 (.65)\\ 
 \hline
 & M2b & 5ep. & 45.01 (.84) & 44.18 (.62) & 44.30 (.96) & 43.33 (.33) & \textbf{43.08 (.17)} & \textbf{43.08 (.17)} & 44.92 (.89)\\ 
 & M2b & 20ep. & 42.50 (.48) & 43.79 (.25) & 43.60 (.64) & 42.72 (.33) & \textbf{42.12 (.15)} & 42.50 (.29) & 43.23 (.27)\\ 
 \hline
 & M4b & 5ep. & 27.74 (.48) & 34.93 (.96) & 28.50 (.68) & \textbf{25.60 (.29)} & 28.61 (.50) & 28.61 (.50) & 29.61 (.99)\\ 
 & M4b & 20ep. & 27.45 (.39) & 29.15 (.67) & 27.84 (.59) & \textbf{25.30 (.18)} & 26.35 (.64) & 25.93 (.64) & 27.94 (.22)
\\ \hline 
\end{tabular}\\}}
\caption{Best performances (test error over all tested parameter settings) at 5 and 20 epochs of \NAG, \Adagrad, \adam, \adameve, \eve\ and \XX\ (average and standard deviation on 5 runs). For each line, the best results are in bold, and results less than 1 std.dev. away are in italics.}
\label{tab:best}
\end{table}

\begin{figure}
\centering
\includegraphics[width=0.8\textwidth]{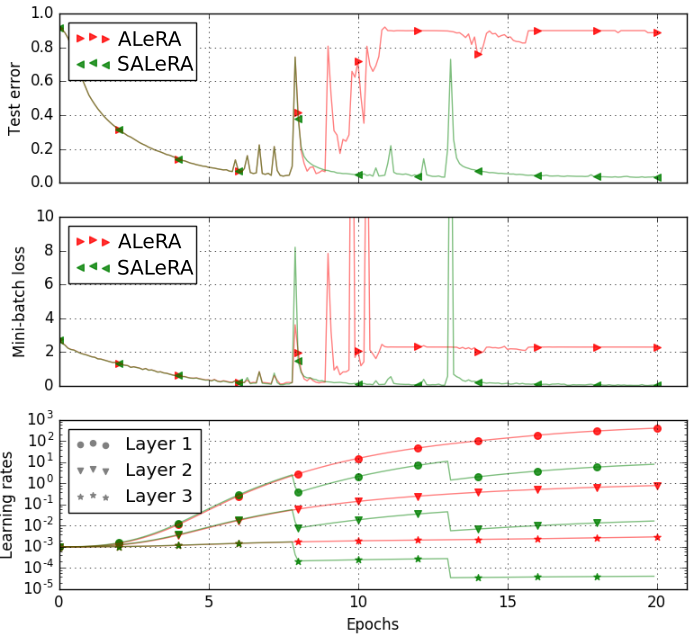} 
\caption{Comparison, on a representative run on MNIST M2 (with same random seed), of \eve\ and \XX: test error (top row), minibatch error (middle row) and learning rates (bottom row). A first catastrophic episode is met around epoch 8: \XX\ reacts by dividing the learning rate for all three layers. Note that the further catastrophic events met by \XX\ are more rare and less severe than for \eve: \XX\ eventually yields a better training loss and a considerably better test error than \eve. Better seen in color.}
\label{fig:lossAndRates}
\end{figure}

\begin{table}[htbp]
{\footnotesize
\begin{tabular}{l | l | l || c c c c c c |} \hline
&  & & \NAG & \Adam & \adameve &  \eve & \XX & \pXX \\ \hline
MNIST & M0 & 5ep. & \textit{7.91 (.14)} & \textit{7.87 (.25)} & \textbf{7.86 (.20)} & \textit{7.96 (.13)} & \textit{7.96 (.13)} & \textit{8.06 (.10)}\\ 
 & & 20ep. & 7.59 (.07) & 7.45 (.08) & \textbf{7.34 (.08)} & 7.53 (.10) & 7.53 (.06) & 7.60 (.08)\\ 
 \hline
 & M2 & 5ep. & 1.95 (.17) & 2.30 (.17) & 2.04 (.15) & \textit{1.89 (.07)} & \textbf{1.87 (.05)} & 1.94 (.13)\\ 
 &  & 20ep. & \textit{1.64 (.06)} & \textbf{1.59 (.06)} & 1.80 (.45) & 1.68 (.04) & 1.68 (.05) &\textit{ 1.63 (.05)}\\ 
 \hline
 & M2b & 5ep. & 1.85 (.05) & 1.93 (.09) & 1.99 (.11) & \textbf{1.71 (.11)} & \textit{1.72 (.04)} & \textit{1.80 (.08)}\\ 
 &  & 20ep. & \textbf{1.47 (.10)} & 1.62 (.09) & 1.82 (.26) & \textit{1.59 (.05)} & \textit{1.59 (.05)} & \textit{1.52 (.02)}\\ 
 \hline
 & M4b & 5ep. & \textbf{.85 (.08)} & 1.03 (.08) & 1.08 (.13) & .99 (.11) & .99 (.14) & .94 (.09)\\ 
 &  & 20ep. & .76 (.12) & .80 (.08) & .91 (.08) & .83 (.01) & .83 (.01) & \textbf{.63 (.11)}\\ 
 \hline
 \hline
CIFAR & M0 & 5ep. & \textit{60.73 (.44)} & \textit{60.74 (.39)} & \textit{60.55 (.49)} & \textit{60.46 (.81)} & \textbf{60.43 (.42)} & 61.71 (.76)\\ 
 & & 20ep. & 59.73 (.19) & 60.08 (.22) & \textit{59.34 (.24)} & \textbf{59.31 (.11)} & \textit{59.37 (.10)} & 59.79 (.57)\\ 
 \hline
 & M2 & 5ep. & \textit{46.62 (.12)} & \textit{45.71 (.54)} & \textit{45.15 (.15)} & \textbf{44.84 (17.52)} & \textit{45.01 (.14)} & \textit{46.08 (.90)}\\ 
 &  & 20ep. & 45.08 (.32) & 44.44 (.61) & 44.30 (.42) &\textit{ 43.92 (1.43)} & \textbf{43.68 (.25)} & 44.62 (.56)\\ 
 \hline
 & M2b & 5ep. & 45.48 (.67) & 44.90 (.50) &\textbf{ 43.33 (.33)} & 44.15 (.29) & 44.49 (.18) & 45.45 (.57)\\ 
 & & 20ep. & 43.86 (.52) & 43.60 (.64) & 43.11 (.35) & \textbf{42.70 (.31)} &\textbf{ 42.70 (.31)} & 43.51 (.80)\\ \hline
 & M4b & 5ep. & \textbf{27.74 (.48)} & 29.34 (.83) & \textit{28.15 (.21) }& 28.61 (.50) & 28.61 (.50) & 30.09 (.73)\\ 
 & & 20ep. & \textit{27.70 (.76) }& 28.00 (.72) & \textbf{27.46 (.54)} & \textit{27.60 (.26)} & \textit{27.60 (.26)} & 28.05 (.43)\\ 
\hline 
\end{tabular}\\}
\caption{Best performance (test error) for the robust algorithm-parameter settings (over all run-parameters settings for $\eta_0$ and batch size): momentum $.9$ for \NAG, $\beta1 = .8$ and $\beta2 = .9999$ for \Adam, $\alpha = .001$ and $C = 3.10^{-6}$ for \adameve\, $\alpha = .01$ and $C = 3.10^{-6}$ for \eve\ and \XX, and $\alpha = .1$ and $C = 3.10^{-8}$ for \pXX.}
\label{tab:robust}
\end{table}
\paragraph{Learning performances}
The experimental evidence (Table \ref{tab:best}) shows that \adameve\ quite often slightly but statistically significantly improves on \adam. A possible explanation is that \adameve\ has a  more flexible adjustment of the learning rate than \adam\ (possibly increasing $\eta$ by a few orders of magnitude). In many cases, \eve\ and \XX\ yield similar results; indeed, whenever \eve\ does not meet catastrophic episodes, \eve\ and \XX\ have the same behaviors. A representative run, where \eve\ and \XX\ undergo catastrophic episodes is depicted on Fig. \ref{fig:lossAndRates} (both runs with same random seed). \eve\ faces a series of catastrophic episodes, where the training error reaches up to 80\%. It eventually stabilizes itself  with a medium training loss, {\em but with test error above 80\%}. In the meanwhile, \XX\ reacts upon the first catastrophic episode around epoch 8, by halving the learning rate on each layer. It faces a further catastrophic episode around epoch 13, and halves the learning rates again. Overall, it faces less frequent and less severe (in terms of train loss and test error deteriorations) accidents. Eventually, \XX\ recovers acceptable train and test errors. 

It is interesting to note that the learning rates on Fig.~\ref{fig:lossAndRates} are constantly increasing, in contradiction with common knowledge. In practice, the learning rate behavior depends on the dataset, the neural architecture and the seed, and can be very diverse (constant decrease, constant increase, or most of the time, an increase followed by a decrease). The diverse learning rate behavior is viewed as an original feature of the proposed approach, made possible by the ability to detect, and recover from, catastrophic explosions of the training loss.

\begin{figure}
\caption{Comparative behaviors on CIFAR-10, model M4 (best configuration for each algorithm, see text). Left: test error. Middle: test loss. Right: mini-batch error. Better seen in color.}
\includegraphics[width=\textwidth]{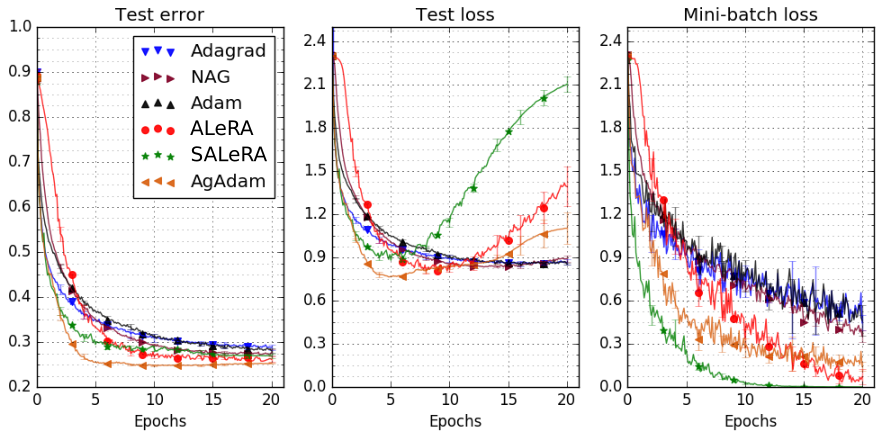} 
\label{fig:convergencePlotsCifar}
\end{figure}
The actual behavior of all algorithms is depicted for CIFAR-10, model M4 on Fig.  \ref{fig:convergencePlotsCifar}, with \adameve, \eve\ and \XX\ respectively getting first, second and third rank in terms of test error at 20 epochs. 
In terms of optimization {\em per se}, \XX\ (respectively \eve) reaches a training error close to 0  at epoch 10 (resp. epoch 20) whereas \adameve\ reaches a plateau after epoch 15. In the meanwhile, {\em the test error decreases and the test loss increases} for all three algorithms. A tentative interpretation for this fact is that the neural net yields more crisp output, close to 0 or 1; this does not change the error while increasing the loss. This result suggests several perspectives for further work (Section \ref{sec:discu}).

In order to determine to which extent the best results in Table~\ref{tab:best} depend on the hyper-parameter settings, and define best configurations for the considered benchmarks, a sensitivity analysis was performed, comparing the results on all models and all epochs for each setting, and choosing the one with lowest sum of ranks. The results for these robust settings are displayed table \ref{tab:robust} (as \Adagrad\ has no hyper-parameter, it is not mentioned there). The proposed hyper-parameters for \eve\ and \XX\ are found to be $\alpha = .01$ and $C = 3.10^{-6}$. Interestingly enough, we find that the optimal hyper-parameters for \Adam\ are $\beta1 = .8$ and $\beta2 = .9999$ instead of $\beta1 = .9$ and $\beta2 = .999$ as is suggested in the original paper~\citep{adam}.
The proposed approaches still show some advantage over \Adam\ and \NAG, though they seem more sensitive to their parameter tuning. It is left for further work to derive precise recommendations depending on model characteristics.

\paragraph{Catastrophe management performances}
Let us define a failed run as a run which attains more than 80\% test error after 20 epochs. \eve~is observed to have 18.3\% failed runs over the parameter range as defined in~\ref{sec:parameters}. Our catastrophe management scheme makes it possible for \XX~to avoid approximately 40\% of these failures, reaching a rate of failure of 11.7\% on the same parameter range.

It would of course be possible to further diminish the failure rate of \XX by setting the PH alarm threshold $\Delta$ to ${\cal L}(\theta, \mbox{first mini-batch})/\lambda$ with $\lambda = 100$ (or $1,000$), rather than $10$ as used in all experiments reported above. However, this would potentially interfere with learning rate adaptation by triggering learning rate halvings even when there is no serious alert to be made, thus preventing it to be as bold as possible. Indeed, setting $\lambda = 100$ causes a decline of more than half of \XX~ performances (data not shown), even though it manages to approximately halve the number of failed runs. Furthermore, setting $\lambda=1~000$ does not further diminish the failure rate, therefore only harming \XX~learning performances. On the other hand, setting $\lambda=1$ does not significantly improve the best performances, but has a higher failure rate.

One tenth of the initial loss is therefore a very good balance between the aggressive learning rate adaptation scheme and its braking counterpart, at least on these datasets and architectures.



\section{Discussion}\label{sec:discu}
The first proposed contribution relies on the comparison of the gradient momentum with a fixed reference. It is meant to estimate the overall correlation among the sequence of gradients, which can be thought of as a signal-to-noise ratio in the process generated from the current solution, the objective and the successive mini-batches. Depending on this ratio, the process can be accelerated or slowed down. The \eve\ procedure, which implements this idea, proves to be significantly able to increase and decrease the learning rate. Furthermore, this process can be plugged on \adam, with a performance improvement on average. 

The price to pay for this flexibility is that it increases the risk of catastrophic episodes, with instant rocketing of the training loss and gradient norm. The proposed approach relies on the conjecture that catastrophic episodes can be rigorously observed and detected. A second conjecture is that a neural net optimizer is almost doomed to face such episodes along the optimization process. These events are mostly detrimental to optimization: before the run, one often chooses small learning rates (and thus slow convergence) to prevent them, and during the run, they are mostly not recovered from. 
Based on these two conjectures, the second contribution of the paper is  an agnostic and principled way to detect and address such episodes. The detection relies on the Page-Hinkley change point detection test. As soon as an event is detected, learning rates are halved and the previous solution recovered. 

A short-term perspective for further research is to apply the proposed approach to recurrent neural networks, and to consider more complex datasets. Another perspective is to replace the halving trick by approximating the line search, e.g. by exploiting the gaps between the actual momentum and the reference one, for several values of the momentum weight factor. A third perspective regards the adaptation of the PH detection threshold $\Delta$ during learning. In some runs, the PH test is triggered over and over, resulting in a very small learning rate preventing any further improvement. The goal is to adapt $\Delta$ when the PH mechanism is re-initialized (line \ref{reinitialisation} of Algorithm \ref{alg.mainLoop}) from the current loss values. Another perspective is to apply \XX\ to 

\subsubsection*{Acknowledgments}
We heartfully thank Steve Altschuler and Lani Wu for making this work possible, supporting it, as well as for very insightful discussions. We also thank Yann Ollivier and Sigurd Angenent for other insightful discussions, and the  anonymous reviewers of a preliminary version of this paper for their accurate and constructive comments.
\bibliographystyle{abbrvnat}
\bibliography{bibliography}
\clearpage
\section*{Appendix A: Derivation of formula \ref{random_mean_t}}
\label{sec:derivationformula}
Let us define r$_0$ = 0, and for $t>0$
\begin{align*}
\RWGt &= \alpha u_t + (1-\alpha) \RWGtm
\end{align*}

with $u_t$ a unit vector uniformly drawn in $\RR^d$. We have by recurrence for $t>0$
\begin{align*}
\RWGt &= \alpha \sum_{l=1}^t (1-\alpha)^{(t-l)} u_l
\end{align*}

Thus
\begin{align}\label{eqq}
\NRGWt &= \alpha^2 \left( \sum_{l=1}^t (1-\alpha)^{2(t-l)} \|u_l\|^2 + \sum_{l\neq k} (1-\alpha)^{2t-l-k} <u_l, u_k> \right)
\end{align}

We are dealing with unit vectors which are uniformly drawn. Thus with $\delta$ denoting the Kronecker delta
\begin{align*}
\forall~ l>0~~&  \|u_l\|^2 = 1\\
\forall~ l,k>0~~& \EE (<u_l, u_k>)=\delta_{lk}
\end{align*}

Taking the expectation in~\ref{eqq} we have
\begin{align}
\EE(\NRGWt) = \alpha^2 \sum_{l=1}^t (1-\alpha)^{2(t-l)}
\end{align}
that is,

\begin{align}
\EE(\NRGWt)& = \alpha^2 \dfrac{1-(1-\alpha)^{2t}}{1-(1-\alpha)^{2}}\\
& = \dfrac{\alpha}{2-\alpha} \left(1-(1-\alpha)^{2t}\right)
\end{align}

The derivation of formula~\ref{random_var_t} is similar.

\clearpage

\section*{Appendix B: Safe Parameter-wise Agnostic LEarning Rate Adaptation}
\label{paramalgo}
Algorithm \ref{pAlgo} gives the detailed implementation of the parameter-wise version of both \eve\ (white lines) and \XX\ (plus greyed lines), as briefly described in Section \ref{pagma}.
In line \ref{cwise}, $\odot$ denotes the coordinate-wise multiplication, and  $\mu' = \mu/d$, $\sigma(d)' = \sigma(d)/\sqrt d$. 

\definecolor{shadecolor}{RGB}{200,200,200}

\begin{algorithm}[h!]\label{pAlgo}
\KwIn{~Model with parameter $\theta$ in $\mathbb{R}^d$ and loss function $\cal L$}
\Parameter{~Memory rate $\alpha$, factor $C$ \hfill //algorithm parameters \\
Initial learning rate $\eta_0$, mini-batch ratio $\rho$
\hfill // run parameters
}
\Initialize{~$t=0; p\leftarrow 0; \forall~ i\in~[|1;d|]~ \eta_i \leftarrow \mathbb{1}$; $\mbox{init} (\theta)$ \hfill // initialization \\
\PH{$L,L_{min}, \ell, \bar{\ell}, \leftarrow 0$; 
$\Delta \leftarrow {\cal L}(\theta, \mbox{first mini-batch})/10$
}  \hfill // initialize Page-Hinkley 
}
\While{\text{stopping criterion not met}}{
$MB \leftarrow \mbox{new mini-batch}$ ; $t\leftarrow t+1$ \hfill // perform forward pass \\
$\ell = \rho {\cal L}(\theta, \mbox{MB}) + (1-\rho) \ell$\\
\PH{$\bar{\ell} \leftarrow  (\ell + t \bar{\ell})/(t+1)$} \hfill //  empirical mean of batch losses \\
\PH{$L \leftarrow L + (\ell-\bar{\ell})$} \hfill // cumulated deviations from mean \\ 
\PH{$L_{min} \leftarrow min (L_{min}, L)$} \hfill // lower bound of deviations \\
\uIf{\PH{$L - L_{min} > \Delta$}}{
\PH{$\theta \leftarrow \theta^{(b)}$; $\eta \leftarrow \eta/2$ } \hfill // Page-Hinkley triggered: backtrack \\
\PH{$L, L_{min}, \bar{\ell}, \ell \leftarrow 0$; $t \leftarrow 0$} \hfill // and re-initialize Page-Hinkley 
}
\Else{
\PH{$\theta^{(b)} \leftarrow \theta$} \hfill // save for possible backtracks\\
$g \leftarrow \nabla_{\theta}{\cal L}(\theta, \mbox{MB})$ \hfill // compute gradient with backward pass\\
$p \leftarrow \alpha g/\lVert g \rVert + (1-\alpha) p$ \hfill // exponential moving average of normalized gradients  \\
\For{i=1,d}{ $\eta_i \leftarrow \eta_0 \eta_i \exp \left( C ( p_i^2 - \mu' ) / \sigma(d)' \right)$ \hfill // parameter-wise agnostic learning rate update \\}
$\theta \leftarrow \theta - \eta \odot g$ \hfill // standard parameter update \label{cwise}
}

}
\caption{\pXX : Parameter-wise \eve\ and \PH{Page-Hickley change detection}}
\end{algorithm}

\clearpage
\section*{Appendix C. Analysis of the dividing factor in 1D}
\label{halvingAppendix}

We use here the context and notations of Section \ref{sec:PH}, and make it even simpler by assuming one dimension (the gradient direction does not change), and minimizing the 1D parabola $F(\theta) = \frac{1}{2}a\theta^2$. It is straightforward to show that, independently of the current solution $\theta_t$, the optimal value for the learning rate $\eta$ is $\eta^* = 1/a$, and that the value above which the loss will deteriorate is $\meta = 2 \eta^*$.

Let us assume that the current learning rate is $\eta_t > \meta$, and that the recovery phase of \XX\ will be used to prevent further catastrophic event, with a dividing factor $\zeta > 1$, and let us show that $\zeta= 2$ is the best trade-off between bringing $\eta$ back into $]0,\meta]$ and then reaching the optimal value $\eta^*$, as discussed in Section \ref{sec:PH}. 

After the PH test has been triggered, a first phase brings $\eta_t$ below $\meta$ by successive divisions by $\zeta$. The number of such divisions is $S = \log_\zeta \frac{\eta_t}{\meta} + 1= \log (\frac{\eta_t}{\meta})/ \log(\zeta)$. 

The second phase uses the standard \eve\ procedure to reach $\eta^*$ from this value $\eta_{t+1} \in ]\frac{\meta}{\zeta}, \meta] $. In the 1D context, let us assume a simplified procedure, that updates $\eta$ by multiplying it by some $1-\varepsilon$ if $\eta>\eta^*$ and by $1+\varepsilon$ if $\eta_t<\eta^*$, for some small $\varepsilon$.

Let us consider two cases, depending on whether $\zeta$ is smaller or greater than $2$. We can compute the expectation of the number of iteration of \XX\ that are needed to reach $\eta^*$ from $\eta_{t+1}$.

If $\zeta\le2$, $\frac{\meta}{\zeta} \ge \eta^*$, hence the further standard update phase decreases $\eta$ by multiplying it by $1-\varepsilon$ until becoming less than (and very close to) $\eta^*$. The length of this update phase increases with $\eta_{t+1}$, thus by construction it is minimal for $\zeta = 2$. In the meanwhile, the length of the division phase decreases as $\zeta$ increases; thus the optimal value for $\zeta \in ]1,2]$ is $\zeta = 2$.

If $\zeta \ge 2$, then $\frac{\eta_{t+1}}{\meta} \in ]\frac{1}{\zeta}, 1[$. Let us consider the two intervals $]\frac{1}{\zeta}, \frac{1}{2}[$ and $]\frac{1}{2}, 1[$, and U and T the respective expectations of the number of \eve\ iterations to reach $\eta^*$. The expectation of the total number of iteration is the sum of U and T, weighted by the probability of arriving in the respective intervals, i.e., the lengths of these intervals. 
These weights are hence $\frac{1}{2}\frac{\zeta-2}{\zeta -1}$ for U and $\frac{1}{2}\frac{\zeta}{\zeta -1}$ for T.

The expected value for $T$ is independent of $\zeta$. For $u \in ]\frac{1}{2}, 1[$, and only counting the number of multiplications needed to get close to $\frac{1}{2}$, the number of iterations is $-\frac{\log(2u)}{\log(1-\varepsilon)} \equiv \frac{\log(2u)}{\varepsilon}$. Integrating over $]-\frac{1}{2}, 1[$ gives
\[
T = \frac{1}{\varepsilon}\int_{\frac{1}{2}}^{1} \log(2u) = \frac{1}{\varepsilon} [t \log(t) - t + t \log(2)]_{\frac{1}{2}}^{1} = 
\frac{1}{\varepsilon} (\log(2) - \frac{1}{2})
\]

Similarly, the expected value for $U$ can be computed over the interval $\frac{1}{\zeta}, \frac{1}{2}[$ using the same approach. Only counting again the number of multiplications needed to get close to $\frac{1}{2}$, it comes 

\[
U = \frac{1}{\varepsilon}\int_{\frac{1}{\zeta}}^\frac{1}{2} -\log(2u) = \frac{1}{\varepsilon} [-t \log(t) + t - t \log(2)]_{\frac{1}{\zeta}}^\frac{1}{2} = \frac{1}{\varepsilon} \left( \frac{1}{2} - \frac{\log(\zeta)}{\zeta} - \frac{1-\log(2)}{\zeta}\right)
\]

Let us summarize now the different computational costs involved after a catastrophe has been detected and before the $\eta^*$ grail is reached. The cost of the dividing iterations only involves a forward pass on the current minibatch. Let us denote this cost by $c_F$. On the other hand, the standard \eve\ iterations have a larger cost, involving a forward pass plus a backward pass and the weights update. Let us denote this cost by $c_B$.

We are looking for the value of $\zeta$ that will minimize the total cost of reaching $\eta^*$ after a catastrophic event has been detected, i.e., that minimizes
\[
c_F S + c_B (\frac{1}{2}\frac{\zeta-2}{\zeta -1} U + \frac{1}{2}\frac{\zeta}{\zeta -1} T )
\]
or, equivalently, that minimizes
\begin{equation}
J(\zeta) = \frac{C}{\log(\zeta)} + \frac{1}{2}\frac{\zeta-2}{\zeta -1}  \left( \frac{1}{2} - \frac{\log(\zeta)}{\zeta} - \frac{1-\log(2)}{\zeta} \right) +
\frac{1}{2}\frac{\zeta}{\zeta -1} (\log(2) - \frac{1}{2})
\label{costJ}
\end{equation}
with
\[
C = \frac{c_F}{c_B} \varepsilon \log (\frac{\eta_t}{\meta})
\]

\begin{figure}
\centering
\includegraphics[width=0.6\textwidth,bb=100 220 500 600]{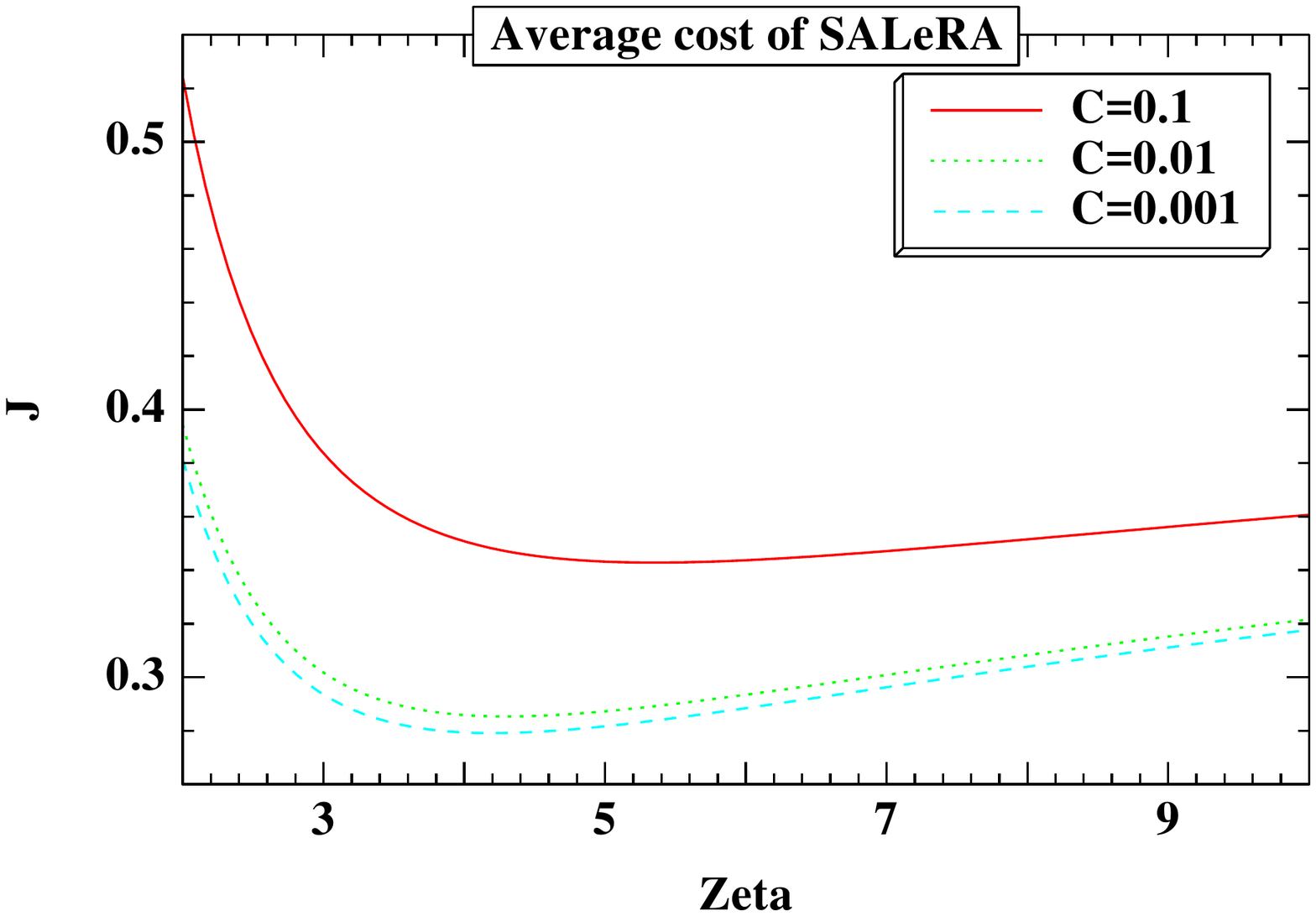}
\caption{Cost function $J$ (Equation \ref{costJ}) for different values of constant C. The plots for smaller values of $C$ are indistinguishable from that of $C=0.001$.}
\label{costAllC}
\end{figure}

It is easy to empirically check (Figure \ref{costAllC}) that $J(\zeta)$ has its global minimum between 3 and 5, that depends on the value of the constant $C$, assumed small here ($C \in ]0, 0.1]$). Then $J$ increases to some asymptotic value. 
However, the value $\zeta = 2$ was initially chosen for historical reasons, by reference to the famed doubling trick frequently used in different areas of Machine Learning. In the light of these results in the simple 1D case, further work will investigate slightly larger values.



\end{document}